\documentclass[10pt,twocolumn,letterpaper]{article}

\usepackage{wacv}              

\usepackage{graphicx}
\usepackage{amsmath}
\usepackage{amssymb}
\usepackage{booktabs}

\usepackage[utf8]{inputenc} 
\usepackage[T1]{fontenc}    
\usepackage{url}            
\usepackage{booktabs}       
\usepackage{amsfonts}       
\usepackage{nicefrac}       
\usepackage{microtype}      
\usepackage{multirow}

\usepackage[accsupp]{axessibility}
\usepackage{amsmath}
\usepackage{tikz}
\usepackage{array}

\usepackage[pagebackref,breaklinks,colorlinks]{hyperref}

\usepackage[capitalize]{cleveref}
\crefname{section}{Sec.}{Secs.}
\Crefname{section}{Section}{Sections}
\Crefname{table}{Table}{Tables}
\crefname{table}{Tab.}{Tabs.}

\newcommand{\ytComment}[1]{\textcolor{black}{#1}}

\def\wacvPaperID{1139} 
\def\confName{WACV}
\def\confYear{2024}

\begin{document}

\title{Rethink Cross-Modal Fusion in Weakly-Supervised Audio-Visual Video Parsing}


\author{
Yating Xu \qquad Conghui Hu \qquad Gim Hee Lee
\\
Department of Computer Science, National University of Singapore\\
{\tt\small xu.yating@u.nus.edu \qquad conghui@nus.edu.sg \qquad gimhee.lee@nus.edu.sg}
}

\maketitle

\begin{abstract}
Existing works on weakly-supervised audio-visual video parsing adopt hybrid attention network (HAN) as the multi-modal embedding to capture the cross-modal context. It embeds the audio and visual modalities with a shared network, where the cross-attention is performed at the input.
However, such an early fusion method highly entangles the two non-fully correlated modalities and leads to sub-optimal performance in detecting single-modality events.
To deal with this problem, we propose the messenger-guided mid-fusion transformer to reduce the uncorrelated cross-modal context in the fusion. The messengers condense the full cross-modal context into a compact representation to only preserve useful cross-modal information.
Furthermore, due to the fact that microphones capture audio events from all directions, while cameras only record visual events within a restricted field of view, there is a more frequent occurrence of unaligned cross-modal context from audio for visual event predictions. We thus propose cross-audio prediction consistency to suppress the impact of irrelevant audio information on visual event prediction.
Experiments consistently illustrate the superior performance of our framework compared to existing state-of-the-art methods.

\end{abstract}

\section{Introduction}
With the ultimate goal of understanding both audio and visual content in video, multimodal video understanding finds a variety of applications in video retrieval \cite{gabeur2020multi}, video surveillance \cite{wu2020not} \etc 
As video is naturally equipped with both audio and visual signals, many prior works have incorporated audio modality into the analyses and shown its benefits to several emerging visual tasks \cite{gao2020listen, owens2018audio, tian2018audio, evangelopoulos2013multimodal, hao2018integrating}.
Audio-Visual Video Parsing (AVVP) \cite{tian2020unified} is one of the most challenging tasks which aims at classifying and localizing the temporal event segments in the audio and visual streams respectively. 
The task requires the model to fully understand video content in both audio and visual streams while only video-level label is provided, as the fine-grained event labels for the two modalities are labour-intensive to source. Since there exists an intractable barrier to access full supervision, models must resort to a weakly-supervised paradigm by learning from the union of all events in a video without any modality and time indication.

\begin{figure}[t]
\centering
\includegraphics[scale=0.42]{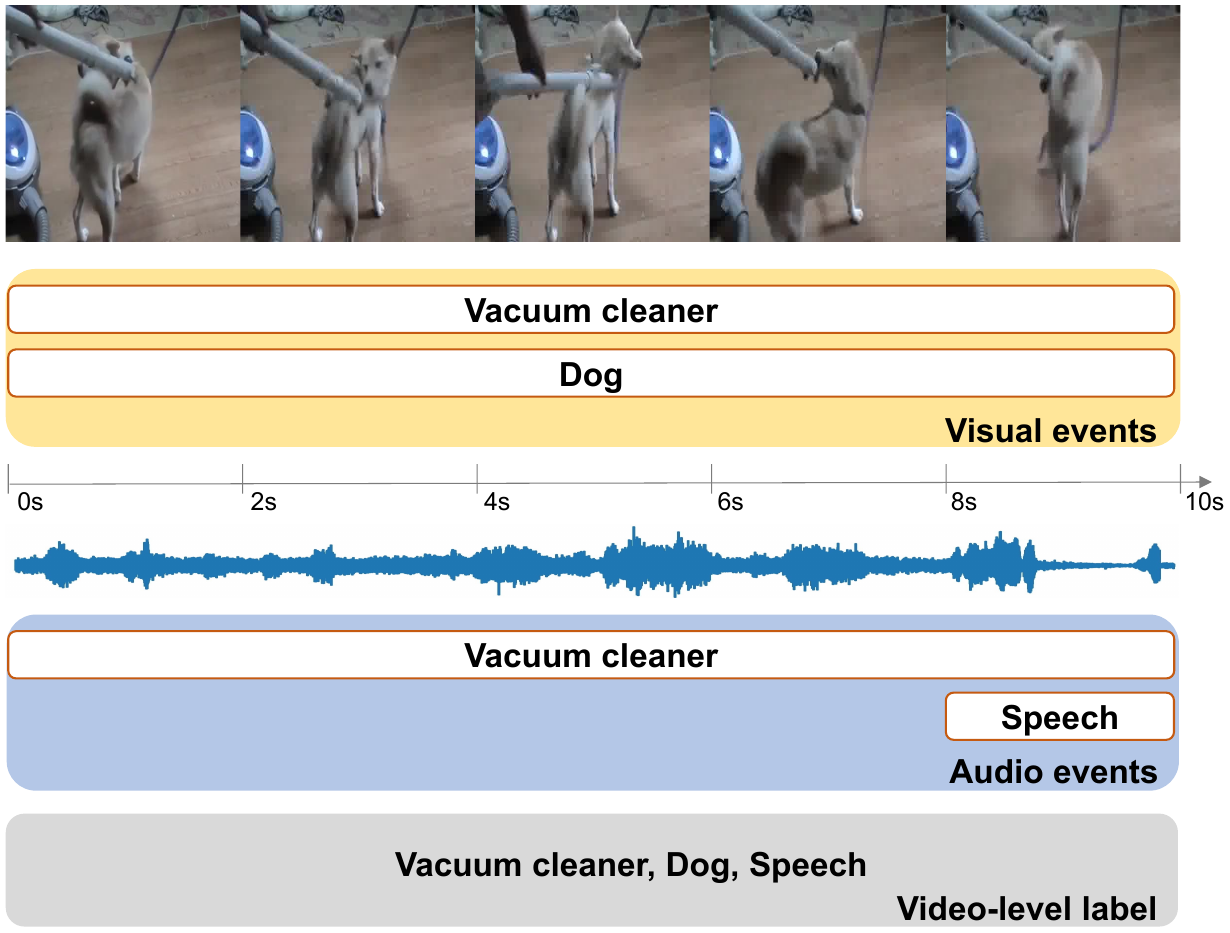}
\caption{Illustration of audio-visual video parsing task. Given a video, it classifies events and detects their temporal locations in the audio and visual streams, respectively. During training, only video-level label is provided.}
\label{teaser}
\end{figure}

\begin{figure}[t]
\centering
\includegraphics[scale=0.65]{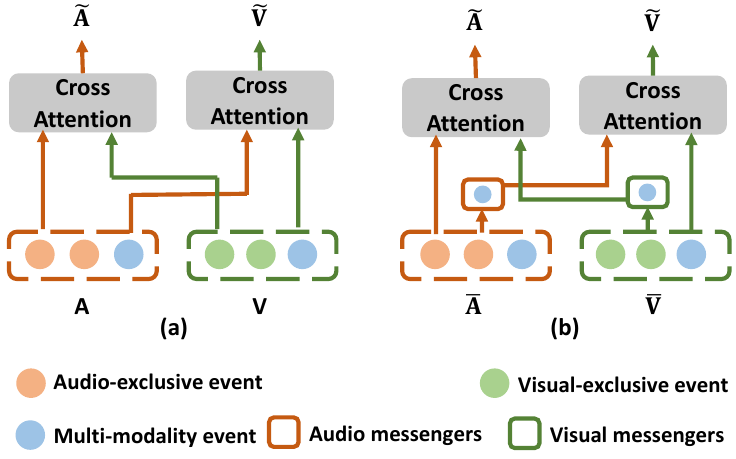}
\caption{Network comparison: (a) HAN embedding and (b) our embedding. A and V denote the pre-extracted audio and visual features, respectively. $\overline{\text{A}}$ and $\overline{\text{V}}$ denotes the self-refined audio and visual features, respectively. $\tilde{\text{A}}$ and $\tilde{\text{V}}$ denote the audio and visual representation after cross-modal fusion, respectively. Audio-exclusive (visual-exclusive) events denotes events only happening on audio (visual) modality.
\ytComment{Each circle represents one event.}
}
\label{network-compare}
\vspace{-3mm}
\end{figure}

To address the challenging problem, the state-of-the-art approaches utilize the correlation between the audio and visual streams to guide the model training.
For example, Hybrid Attention Network (HAN) \cite{tian2020unified} captures the cross-modal context by fusing the pre-extracted audio and visual features directly from the first layer of the network using cross-attention (as illustrated in Fig.~\ref{network-compare}(a)). 
However, this strong entanglement could undesirably mix uncorrelated information from audio and visual streams.
\ytComment{
Real scenes may include occlusion or may be captured by camera of limited field of view, which causes the audio and visual streams not fully correlated.}
For instance, in Fig.~\ref{teaser}, the silent dog and off-camera human speech only appear in one single modality. 
Accordingly, we analyze the performance of HAN in detecting single-modality and multi-modality events as shown in Tab.~\ref{han}. 
Single-modality events denote events \textit{only} happening in audio or visual modality, while multi-modality events refer to events appearing with temporal overlap on audio and visual streams.
The results are averaged F-scores per event.
Compared with the original HAN, the HAN-CA\footnote{We change the input of the cross-attention module to the modality itself so as to keep the model size unchanged.}, when excluding cross-modal fusion, exhibits a significant decrease in predicting multi-modality events. However, the prediction performance for single-modality events experiences an enhancement.
It suggests that strong entanglement with another non-fully correlated modality is harmful in detecting its own exclusive events while the absence of fusion negatively affects the detection of audio-visual events.
As either fully entangled fusion or complete independence of the two modalities can hurt the performance badly, it is imperative to design a better fusion strategy for the two partially correlated modalities.

\begin{table}[t]
\centering
\resizebox{0.8\linewidth}{!}{
\begin{tabular}{c|c|c}
\toprule
Model & Single-modality Event & Multi-modality Event \\ \midrule
HAN   &    44.0                   & \textbf{67.2}                     \\
HAN-CA  &   47.0                  & 58.8                    \\ 
Ours  &     \textbf{50.6}             & 66.1 \\
\bottomrule
\end{tabular}
}
\caption{Analysis of HAN \cite{tian2020unified}. `HAN-CA' denotes HAN without cross-attention. 
Segment-level evaluation is conducted.
}
\vspace{-3mm}
\label{han}
\end{table}

To solve this problem, we propose messenger-guided mid-fusion transformer (MMT) to suppress the uncorrelated information exchange during fusion.
Compared to the early fusion in HAN, the mid-fusion is more flexible in controlling the flow of the cross-modal context. 
It can first aggregate a clearer global understanding of the raw input sequence, which helps identify the useful cross-modal context in the fusion module.
The messengers are the core of MMT, which serves as the compact cross-modal representation during fusion. 
 (Fig.~\ref{network-compare}(b)).
Due to their small capacity, they can help amplify the most relevant cross-modal information that agrees best with the clean labels while suppressing the noisy information that causes disagreement.
Our MMT is able to largely improve the performance of detecting single-modality events while maintaining a relatively high performance of detecting multi-modality events.

We further propose cross-audio prediction consistency (CAPC) to suppress the undesired predictions in the visual stream caused by mismatched audio information. As pointed out by \cite{wu2021exploring}, the ``audible but not visible events'' are more common than ``visible but not audible events'' since the camera only captures the scene of limited view, while the microphones capture events from all directions. Thus, the visual modality is more likely to encounter the non-correlated cross-modal context. 
To alleviate such situations, we introduce CAPC.
Our idea is to allow the visual modality to learn from beyond its paired audio, and induce it to have consistent visual event predictions as learning with its original pair. 
As such, the visual stream learns to only fuse the audio context that is correlated with itself and ignores other unrelated information to maintain the same visual event detection under different audio contexts.

In summary, our contributions are as follows:
\begin{itemize}
\item We propose messenger-guided mid-fusion transformer to reduce the uncorrelated cross-modal context in the audio-visual fusion. The messengers serve as a compact representation to amplify the most relevant cross-modal information under weak supervision.
\item We propose cross-audio prediction consistency to calibrate the visual event prediction under interference from the unmatched audio context. The visual event prediction is forced to remain unchanged when pairing with different audios so as to ignore the irrelevant audio information.
\item We conduct extensive qualitative and quantitative experiments to analyze the effectiveness of our approach. Our method also achieves the state-of-the-art performance on AVVP benchmark.
\end{itemize}

\vspace{0.5em}
\section{Related Work}
\label{related-work}

\subsection{Audio-Visual Representation Learning}
The natural correspondence between audio and visual modalities overcomes the limitations of perception tasks in single modality and introduces a series of new applications. Semantic similarity is the most commonly used audio-visual correlation \cite{arandjelovic2017look, aytar2016soundnet, arandjelovic2018objects,hu2019deep,morgado2021audio}. The shared semantic information in both audio and visual modality is a valuable free-source supervision. SoundNet \cite{aytar2016soundnet} learns sound representation by distilling knowledge from the pre-trained visual models. Morgado \etal \cite{morgado2021audio} propose cross-modal contrastive learning, where negative and positive samples of visual frames are drawn from audio samples and vice-versa. Besides semantic correlation, other works utilize temporal synchronization \cite{owens2018audio, korbar2018cooperative, afouras2020self, shang2021multimodal}, motion correlation \cite{gan2020music, zhao2019sound} and spatial correspondence \cite{yang2020telling, gao2020visualechoes, morgado2020learning}. However, self-supervised learning from natural videos is potentially noisy as the audio and visual modalities are not always correlated. Recently, Morgado \etal \cite{morgado2021robust} propose to learn robust audio-visual representation by correcting the false alignment in the contrastive loss. Our task shares similar motivation with \cite{morgado2021robust} as some events only happen in single modality leaving no audio-visual correspondence.

\subsection{Audio-Visual Video Parsing}
Early works \cite{tian2018audio, lin2019dual} only detect events that is both audible and visible. Based on the strong assumption that the audio and visual information are aligned at each time step, \cite{tian2018audio, lin2019dual} fuse the audio and visual features at the same time step.
However, the events happening on the two modalities are not always the same since the audio and vision are inherently different sensors. To fully understand the content in the multimodal videos, Tian \etal \cite{tian2020unified} introduced the task of audio-visual video parsing (AVVP). It classifies and localizes all the events happening on the audio and visual streams in a weakly-supervised manner. They design a hybrid attention network (HAN) to capture the uni-modal and cross-modal temporal contexts simultaneously. The audio and visual features are fused at the start of the network, where the self-attention and cross-attention are performed in parallel. 
Since then, HAN serves as the state-of-the-art audio-visual embedding and is widely adopted in follow-up works. MA \cite{wu2021exploring} generates reliable event labels for each modality by exchanging the audio and visual tracks of a training video with another unrelated video. JoMoLD \cite{cheng2022joint} leverage audio and visual loss patterns to remove modality-specific noisy labels for each modality. Lin \etal \cite{lin2021exploring} explore the cross-modality co-occurrence and shared cross-modality semantics across-videos. Although HAN embedding shows promising performance, the full entanglement of two non-fully correlated modalities is not ideal for the task of AVVP.
Therefore, we propose messenger-guided mid-fusion transformer and cross-audio prediction consistency to reduce the uncorrelated cross-modal context in the audio-visual fusion.

\subsection{Multimodal Transformer}
The attention module in the transformer \cite{vaswani2017attention} is effective in capturing the global context among the input tokens, and is widely adopted in the multimodal task \cite{lu2019vilbert, gabeur2020multi, girdhar2022omnivore, jaegle2021perceiver, nagrani2021attention}. Gabeur \etal \cite{gabeur2020multi} use transformer to capture cross-modal cues and temporal information in the video. OMNIVORE \cite{girdhar2022omnivore} proposes a modality-agnostic visual model that can perform classification on image, video, and single-view 3D modalities using the same shared model parameters. Perceiver \cite{jaegle2021perceiver} and MBT \cite{nagrani2021attention} address the high computation cost of the multimodal transformer by using a small set of fusion tokens as the attention bottleneck to iteratively distill the uni-modal inputs. Despite our messengers also serve as the attention bottleneck, it is used to suppress learning from noisy labels. Moreover, the messengers are more effective in the small multimodal models. The MBT and Perceiver initialize the fusion tokens randomly and require multiple times updates with the uni-modal inputs for it to carry meaningful cross-modal information, which is not applicable for the model with small number of encoder layers. In contrast, our messenger is directly derived from the global representation of each modality so that it is already informative without multiple times of updates. 


\begin{figure*}[t]
\centering
\includegraphics[scale=0.62]{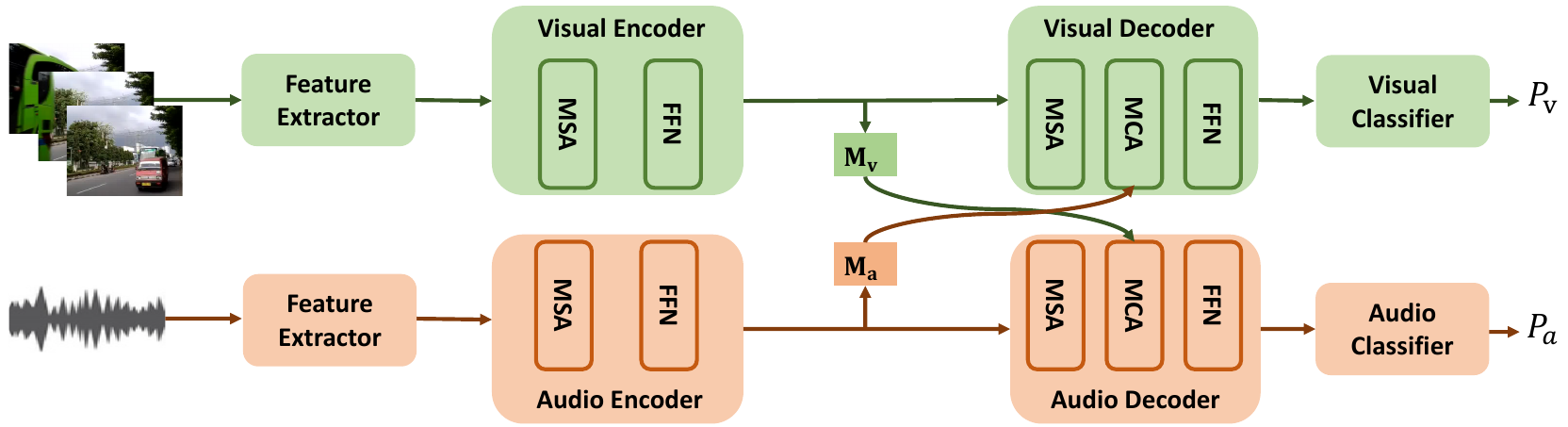}
\caption{The architecture of the messenger-guided mid-fusion transformer. Visual Encoder and Audio Encoder denote the transformer encoders for the visual and audio, respectively. Visual Decoder and Audio Decoder denote the transformer decoders for the visual and audio, respectively.}
\label{framework}
\end{figure*}

\section{Method}
\label{method}
Let us denote a video with $T$ non-overlapping segments as $\{\text{V}_t, \text{A}_t\}_{t=1}^T$, where $\text{V}_t$ and $\text{A}_t$ are the visual and audio clip at the $t$-th segment, respectively. 
The corresponding label for visual event, audio event and audio-visual event at the $t$-th segment is denoted as $y_{t}^\text{v} \in \{0,1\}^{C}$, $y_{t}^\text{a} \in \{0,1\}^{C}$ and $y_{t}^\text{av}\in \{0,1\}^{C}$, respectively. $C$ is the total number of classes in the dataset and $y_{t}^\text{av} = y_t^\text{v} \times y_t^\text{a}$. An event is considered as an audio-visual event only if it occurs in both modalities. Note that more than one event can happen in each segment.  Grouping all the segment-level labels together, we obtain the video-level label $Y = \{y_t^\text{v} \cup y_t^\text{a}\}_{t=1}^T \in \{0,1\}^C$. The goal of audio-visual video parsing is to detect all the visual, audio and audio-visual events in the video. The training of AVVP is conducted in weak supervision, where only video-level labels $Y$ are provided. 

In the following sections, we first introduce messenger-guided mid-fusion transformer as the new multi-modal embedding for AVVP, and then the novel idea of cross-audio prediction consistency to reduce the negative interference of the unmatched audio context to the visual stream.

\subsection{Messenger-guided Mid-Fusion Transformer}
Fig.~\ref{framework} shows our proposed messenger-guided mid-fusion transformer. We instantiate the self-attention and cross-attention layers with transformers \cite{vaswani2017attention} as it shows excellent performance in the uni-modal \cite{dosovitskiy2020image, arnab2021vivit} and multimodal  \cite{gabeur2020multi, seo2021look, girdhar2022omnivore} tasks with just the attention mechanism. The pre-trained visual and audio feature extractors extract segment-level visual features $\{f_t^\text{v}\}_{t=1}^T$ and audio features $\{f_t^\text{a}\}_{t=1}^T$, respectively. The $\{f_t^\text{v}\}_{t=1}^T$ and $\{f_t^\text{a}\}_{t=1}^T$ are the input to the multimodal embedding, where the uni-modal and cross-modal context are modeled sequentially. 
Instead of directly feeding the full cross-modal context to the fusion, we summarize it into compact messengers. Finally, the outputs of the last layer of the decoders are sent into the classifiers to detect segment-level events for each modality.   

\paragraph{Uni-modal Context Refinement.} 
We model the uni-modal context with transformer encoders, where the self-attention is capable of aggregating the temporal context for a better global understanding of the raw input sequence. For brevity, we only illustrate the working flow of the visual modality since the visual and audio branches work symmetrically. The pre-extracted visual features are first converted into 1-D tokens $S_\text{v} \in \mathbb{R}^{T\times d}$ with feature dimension $d$ as follow:
\begin{equation}
        S_\text{v} =\left[f_1^\text{v} \mathbf{W}_\text{v}^{\text{enc}}, f_2^\text{v} \mathbf{W}_\text{v}^{\text{enc}}, \ldots, f_\text{T}^\text{v} \mathbf{W}_\text{v}^{\text{enc}}\right]+\mathbf{PE},
\end{equation}
where $\mathbf{W}_\text{v}^\text{enc}$ is the linear projection layer that projects pre-extracted features to \textit{d} dimension and $\mathbf{PE}$ is the position embedding. Then, the tokens are sent into a $L$-layer transformer encoder. We adopt the original architecture of transformer \cite{vaswani2017attention}. Each layer consists of a multi-headed self-attention ($\operatorname{MSA}$) and a position-wise fully connected feed-forward network ($\operatorname{FFN}$):    
\begin{equation}
\begin{aligned}
    \tilde{S}_\text{v}^l &= \operatorname{LN}\left(\operatorname{MSA}\left(S_\text{v}^l\right) + S_\text{v}^l\right), \\   
    S_\text{v}^{l+1} &= \operatorname{LN}\left(\operatorname{FFN}\left(\tilde{S}_\text{v}^l\right) + \tilde{S}_\text{v}^l\right),
    \end{aligned}
\end{equation}
where $\operatorname{LN}$ denotes layer normalization and $S_\text{v}^l$ is the input tokens at the $l$-th layer.

\begin{figure*}[t]
\centering
\includegraphics[scale=0.7]{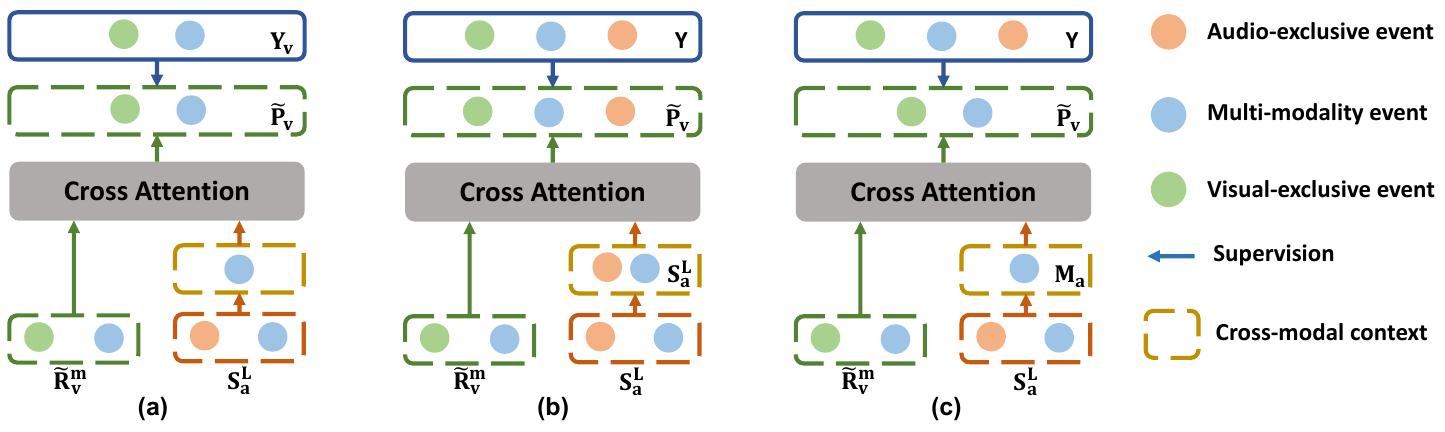}
\caption{Comparison of different fusion methods. Each circle represents one event. We only illustrate the cross-modal fusion at the visual branch for brevity. (a) shows an oracle setting, where the visual label $Y_\text{v}$ is available. In this case, the model only learns from the relevant audio context. (b) shows the fusion with full audio context. (c) shows the fusion with messengers.
}
\label{compare-msg}
\end{figure*}

\paragraph{Cross-modal Context Fusion with Messengers.}

We model the cross-modal context using $M$-layer transformer decoders, where each layer consists of the multi-headed self-attention ($\operatorname{MSA}$), multi-headed cross-attention ($\operatorname{MCA}$), and the position-wise feed-forward network ($\operatorname{FFN}$). The working flow inside the $m$-th layer of the decoder is as follows:
\begin{equation}
    \begin{aligned}
    \tilde{R}_\text{v}^m &= \operatorname{LN}\left(\operatorname{MSA}\left(R_\text{v}^m\right) + R_\text{v}^m\right), \\
    \hat{R}_\text{v}^m &= \operatorname{LN}\left(\operatorname{MCA}\left(\tilde{R}_\text{v}^m, S_\text{a}^L\right) + \tilde{R}_\text{v}^m\right), \\
    R_\text{v}^{m+1} &= \operatorname{LN}\left(\operatorname{FFN}\left(\hat{R}_\text{v}^m\right) + \hat{R}_\text{v}^m\right),
    \end{aligned}
\label{fuse without messenger}
\end{equation}
where $\operatorname{MCA}(\cdot)$ performs cross-modal fusion between the visual feature $\tilde{R}_\text{v}^m$ (query) and the audio context $S_\text{a}^L$ (key and value).

However, providing full cross-modal context is not ideal when the two modalities are not fully correlated, and is even worse when the supervision is noisy. The video-level label $Y$ is the union of the audio and visual events, which can introduce noise when supervising each modality, \ie an event that is present only in one modality becomes a noisy label for the other modality. As shown in Fig.~\ref{compare-msg}(b), with connection with full audio context, the model is assured that the audio-exclusive event truly happens in the visual stream guided by the noisy supervision Y. Consequently, its generalization ability is severely affected.
To this end, we create a fusion bottleneck $M_\text{a}$ as shown in Fig.~\ref{compare-msg}(c) so that they can suppress the irrelevant audio context, and the model is less likely to overfit to the noisy label $Y$.


Specifically, we condense the full cross-modal context into the compact representation $M_\text{v}$ as follows:
\begin{equation}
   M_\text{v} = \operatorname{Tanh}\left(\operatorname{Pool}\left(S_\text{v}^L \mathbf{W}_\text{v}^\text{msg}; n_\text{v}\right)\right),
\end{equation}
where $\mathbf{W}_\text{v}^\text{msg} \in \mathbb{R}^{d \times d}$ is the linear projection layer, $\operatorname{Pool}(\cdot)$ is the average pooling along the temporal dimension with a target length of $n_\text{v}$ and $\operatorname{Tanh}(\cdot)$ is the activation function. $M_\text{v} \in \mathbb{R}^{n_v \times d}$ has limited capacity in storing information compared to the full cross-modal context $S_v^L \in \mathbb{R}^{T \times d}$, where T is much larger than $n_v$. Consequently, it creates an attention bottleneck that gives priority to the most relevant cross-modal context that fits the clean labels.
The compact audio context $M_\text{a} \in \mathbb{R}^{n_a \times d}$ at the audio stream is obtained in a similar way. We name it messenger as it represents its source modality as the direct input in the cross-modal fusion as follows: 
\begin{equation}
    \hat{R}_\text{v}^m = \operatorname{MCA}\left(\operatorname{LN}\left(\tilde{R}_\text{v}^m, M_\text{a}\right)\right) + \tilde{R}_\text{v}^m, \\
    \label{eqn_fusion_messenger}
\end{equation}
where the audio messengers $M_\text{a}$ replace the full cross-modal context $S_\text{a}^L$.

\paragraph{Classification.} $R_\text{v}^M$ is considered as the final visual representation and is sent into the modality-specific classifiers for segment-level event prediction $P_\text{v} \in \mathbb{R}^{T \times C}$ as follow:
\begin{equation}
    P_\text{v} = \operatorname{Sigmoid}\left(R_\text{v}^{M} \mathbf{W}_\text{v}^{\text{cls}}\right) ,
\end{equation}
where $\mathbf{W}_\text{v}^{\text{cls}} \in \mathbb{R}^{d \times c}$ is the classifier weights. During training, we aggregate $P_\text{v}$ into video-level predictions $\tilde{P}_\text{v} \in \mathbb{R}^{C}$ via soft pooling similar to \cite{tian2020unified} since only video-level label $Y \in \{0,1\}^C$ is provided. The audio prediction $P_\text{a}$ and $\tilde{P}_\text{a}$ are obtained in a similar way. We also combine $\tilde{P}_\text{v}$ and $\tilde{P}_\text{a}$ into a modal-agnostic video-level prediction $\tilde{P}_{\text{video}} \in \mathbb{R}^C$. In total, we have three classification losses:
\begin{equation}
    \mathcal{L}_\text{cls} = \operatorname{CE}\left(\tilde{P}_\text{v}, Y_\text{v}\right) + \operatorname{CE}\left(\tilde{P}_\text{a}, Y_\text{a}\right) +\operatorname{CE}\left(\tilde{P}_{\text{video}}, Y\right),
\end{equation}
where $\operatorname{CE}$ denotes the binary cross-entropy loss. We set $Y_\text{v}=Y_\text{a}=Y$  when only the video-level label is available.

\begin{table*}[t]
\centering
\resizebox{\linewidth}{!}{
\begin{tabular}{p{2.5cm}<{\centering}|p{1.2cm}<{\centering}p{1.2cm}<{\centering}|p{1.2cm}<{\centering}p{1.2cm}<{\centering}|p{1.2cm}<{\centering}p{1.2cm}<{\centering}|p{1.2cm}<{\centering}p{1.2cm}<{\centering}|p{1.2cm}<{\centering}p{1.2cm}<{\centering}}
\toprule
\multirow{2}{*}{Method} & \multicolumn{2}{c|}{Audio} & \multicolumn{2}{c|}{Visual} & \multicolumn{2}{c|}{Audio-Visual} & \multicolumn{2}{c|}{Type@AV} & \multicolumn{2}{c}{Event@AV} \\
 & Seg. & Event & Seg. & Event & Seg. & Event & Seg. & Event & Seg. & Event \\
 \midrule
AVE \cite{tian2018audio} & 47.2 & 40.4 & 37.1 & 34.7 & 35.4 & 31.6 & 39.9 & 35.5 & 41.6 & 36.5 \\
AVSDN \cite{lin2019dual} & 47.8 & 34.1 & 52.0 & 46.3 & 37.1 & 26.5 & 45.7 & 35.6 & 50.8 & 37.7 \\
HAN  \cite{tian2020unified} & 60.1 & 51.3 & 52.9 & 48.9 & 48.9 & 43.0 & 54.0 & 47.7 & 55.4 & 48.0 \\
HAN $\dag$ \cite{tian2020unified} & 59.8 & 52.1 & 57.5 & 54.4 & 52.6 & 45.8 & 56.6 & 50.8 & 56.6 & 49.4 \\
MA \cite{wu2021exploring} & 60.3 & 53.6 & 60.0 & 56.4 & 55.1 & 49.0 & 58.9 & 53.0 & 57.9 & 50.6 \\
Lin \etal \cite{lin2021exploring} & 60.8 & 53.8 & 63.5 & 58.9 & 57.0 & 49.5 & 60.5 & 54.0 & 59.5 & 52.1 \\
JoMoLD \cite{cheng2022joint} &61.3 & \textbf{53.9} & 63.8& 59.9 & 57.2 & 49.6 & 60.8 & 54.5 & 59.9 & 52.5 \\
Ours & \textbf{61.9} & \textbf{53.9} & \textbf{64.8} & \textbf{61.6} & \textbf{57.6} & \textbf{50.2} & \textbf{61.4} & \textbf{55.2} & \textbf{60.9} & \textbf{53.1} \\
\bottomrule
\end{tabular}
}
\vspace{-3mm}
\caption{Comparison with the state-of-the-art methods of audio-visual video parsing on the LLP test dataset. `Audio', `Visual' and `Audio-Visual' denotes audio event, visual event and audio-visual event detection, respectively. 
\ytComment{
Note that they are different from the event categories in Tab.~\ref{han} and we illustrate the difference in the Supplementary Material.
}`Seg.' denotes segment-level evaluation and `Event' denotes event-level evaluation. `HAN$\dag$' is the variant of HAN that additionally uses label refinement. The best result is marked in bold.}
\label{compare sota}
\end{table*}

\subsection{Cross-Audio Prediction Consistency}
We further propose cross-audio prediction consistency (CAPC) to suppress the inaccurate visual event prediction arising from unmatched audio information. As analyzed in \cite{wu2021exploring}, audio-exclusive events are more common than visual-exclusive events and thus the visual stream is more likely to be influenced by the non-correlated cross-modal context.
As shown in Fig.~\ref{qualitative-result}, the visual branch confidently detects the audio-exclusive event `Chicken rooster' when only learnt with its paired audio.
To alleviate this problem, we introduce a consistency loss in the visual event prediction by pairing the same visual sequence with different audios. 
Specifically, the visual modality $\text{V}$ is paired with not only its original audio counterpart $\text{A}_{\text{orig}}$, but is also paired with audios that are randomly selected from other videos at each training iteration. We denote the visual prediction from the original pair $(\text{V}, \text{A}_{\text{orig}})$ as $\tilde{P}_\text{v} \in \mathbb{R}^{C}$, and the visual prediction from the $i$-th random pair $(\text{V}, \text{A}_{\text{rand}}^i)$ as $\tilde{P}_\text{v}^{i} \in \mathbb{R}^{C}$. CAPC requires $\tilde{P}_\text{v}^{i}$ to be the same as $\tilde{P}_\text{v}$ as follow:
\begin{equation}
    \mathcal{L}_\text{ccr} = \frac{1}{N} \sum_{i=1}^N \left\|\tilde{P}_\text{v}^{i} - \tilde{P}_{\text{v}}\right\|_{2}^{2},
\end{equation}
where $N$ is the number of random pairs for each visual sequence. 
The cross-attention at the visual stream will learn to 
only grab the useful audio context (\ie audio-visual event) from A and $\text{A}_\text{rand}^{i}$ and ignore the irrelevant information (\ie audio-exclusive event) in order to achieve this prediction consistency.

We notice that there may be a trivial solution to achieve this prediction consistency. The cross-attention totally ignores all the audio context and thus leads to complete independence of the visual prediction from the audio information. However, we show in Tab.~\ref{ablation consistency} that the model does not degenerate to this trivial solution. Instead, CAPC improves the robustness of fusion under non-fully correlated cross-modal context. 

Finally, the total loss of our method is:
\begin{equation}
    \mathcal{L}_\text{total} = \mathcal{L}_\text{cls} + \mu \mathcal{L}_\text{ccr},
\label{full loss}
\end{equation}
where $\mu$ is the hyperparameter to balance the loss terms. 

\subsection{Discussions on CAPC}
\paragraph{Comparison with Consistency Regularization.}
Consistency regularization is widely adopted in semi-supervised learning \cite{sohn2020fixmatch,bachman2014learning,berthelot2019mixmatch}, where the model is required to output the similar prediction when fed perturbed versions of the same image. In contrast to augmenting the modality itself, the CAPC keeps the modality itself intact and only augments its cross-modal context, \ie pairing the visual modality with different audios. By learning consistency under cross-audio augmentation, the fusion robustness is improved.
\vspace{-3mm}
\paragraph{Comparison with Audio-Visual Correspondence.}
Audio-visual pairing correspondence \cite{arandjelovic2018objects, arandjelovic2017look} and audio-visual temporal correspondence \cite{korbar2018cooperative,owens2018audio} are the cross-modal self-supervision in the video. They highlight the audio-visual alignment to learn good audio and visual representation. In contrast, we highlight the audio-visual misalignment to reduce the negative impact of the unmatched audio context to the visual modality.

\section{Experiments}
\label{exp}
\subsection{Experiment Setup}
\label{exp setup}
\paragraph{Dataset.} We conduct experiments on the \textit{Look, Listen and Parse (LLP) Dataset} \cite{tian2020unified}. It contains 11,849 YouTube videos, each is 10-second long. It covers 25 real-life event categories, including human activities, animal activities, music performances, \etc 7,202 video clips are labeled with more than one event category and per video has an average of 1.64 different event categories. We follow the official data splits \cite{tian2020unified}. 10,000 video clips only have video-level labels and are used as training sets. The remaining 1,849 videos that are annotated with audio and visual events and second-wise temporal boundaries are divided into 849 videos as the validation set and 1,000 as the test set. 

\vspace{-5mm}
\paragraph{Evaluation Metrics.} 
We use F-scores as the quantitative evaluation method. We parse the visual, audio and audio-visual events, denoted as `Visual', `Audio' and `Audio-Visual', respectively. 
We use F-scores for segment-level and event-level evaluations. The segment-level metrics evaluate the predictions for each segment independently. The event-level evaluation first concatenates consecutive positive segments as the event proposal and then compares the alignment with the ground-truth event snippet under the mIoU=0.5 as the threshold. Meanwhile, we use `Type@AV' and `Event@AV' for the overall access of the model performance. The `Type@AV' averages the evaluation scores of `Audio', `Visual' and `Audio-Visual. The `Event@AV' considers all the audio and visual events for each video rather than directly averaging results from different event types.

\vspace{-5mm}
\paragraph{Implementation Details.}
Each video is downsampled at 8 fps and divided into 1-second segments. We use both the ResNet-152 \cite{he2016deep} model pre-trained on ImageNet and 18 layer deep R(2+1)D \cite{tran2018closer} model pre-trained on Kinetics-400 to extract visual features. The 2D and 3D features are concatenated as the visual representation for the visual input. For the audio signals, we use the VGGish network \cite{hershey2017cnn} pre-trained on AudioSet \cite{gemmeke2017audio} to extract 128-D features. The feature extractors are fixed during training. For each modality, both the number of encoders L and the number of decoders M are set to 1. The hidden size is set to 512 and the number of attention head is set to 1. The number $n_a$ of audio messengers and the number $n_v$ of visual messengers are set to 1, respectively. 

We train our model in three stages. In the first stage, we optimize our proposed audio-visual embedding with classification loss $\mathcal{L}_\text{cls}$ on the video-level label Y. In the second stage, we calculate the pseudo label \cite{wu2021exploring} for each modality. In the third stage, we re-train our embedding with Eqn.~\ref{full loss} using the pseudo label. $\mu$ is set to 0.5 and N is set to 1. We use Adam optimizer with learning rate $3\times 10^{-4}$ and batch size of 64. We train 40 epochs and decrease the learning rate by $10^{-1}$ every 10 epochs in each training stage. All the experiments are conducted using Pytorch on a NVIDIA GTX 1080 Ti GPU.

\subsection{Comparison with State-of-the-art Results}
We compare our method with state-of-the-art audio-visual event parsing methods of AVE\cite{tian2018audio}, AVSDN \cite{lin2019dual}, HAN \cite{tian2020unified}, MA \cite{wu2021exploring}, Lin \etal \cite{lin2021exploring} and JoMoLD \cite{cheng2022joint}. We report their results from their paper. All the methods are trained using the same pre-extracted features as input. The recent methods, MA \cite{wu2021exploring}, Lin \etal \cite{lin2021exploring} and JoMoLD \cite{cheng2022joint} all adopt the HAN \cite{tian2020unified} as the audio-visual embedding.

Tab.~\ref{compare sota} shows the quantitative comparisons on the LLP dataset \cite{tian2020unified}. Our model constantly outperforms other methods on all the evaluation metrics. Although we remove the entanglement in the early layers, the performance of audio and visual event detection are both improved than all the HAN-based methods. This demonstrates that a compact fusion is better than the fully entangled fusion approach when the audio and visual information are not always correlated.

\begin{table}[t]
\centering
\resizebox{\linewidth}{!}{
\begin{tabular}{cc|c|c|c|c|c}
\toprule
$n_a$ & $n_v$ & Audio & Visual & Audio-Visual & Type@AV & Event@AV \\
\midrule
\multicolumn{2}{c|}{No MSG} & 61.5 & 63.2 & 55.7 & 60.1 & 59.9 \\
\multicolumn{2}{c|}{MBT}  & 60.9 & 64.1 & 56.0  & 60.3  & 59.8 \\
\midrule
5 & 5 & 61.5 & 64.4 & 57.0 & 61.0 & 60.6 \\
3 & 3 & \textbf{62.1} & 63.9 & 56.9 & 61.0 & 60.5 \\
1 & 3 & 61.6 & 64.7 & 56.8 & 61.0 & 60.6 \\
3 & 1 & \textbf{62.1} & 64.6 & 56.8 & 61.2  & \textbf{60.9}  \\
1 & 1 & 61.9 & \textbf{64.8} & \textbf{57.6} & \textbf{61.4} & \textbf{60.9}\\
\bottomrule
\end{tabular}
}
\vspace{-3mm}
\caption{Ablation study of the messengers. 
`No MSG' denotes the model performs cross-modal fusion without messengers. `MBT' replaces the messengers with the fusion bottleneck token \cite{nagrani2021attention}. $n_\text{a}$ and $n_\text{v}$ is the number of audio and visual messengers, respectively, and `$n_\text{a}=1, n_\text{v}=1$' is our final model. Segment-level results are reported.
}
\label{ablation messenger}
\vspace{-3mm}
\end{table}

\begin{figure}[t]
\centering
\includegraphics[scale=0.63]{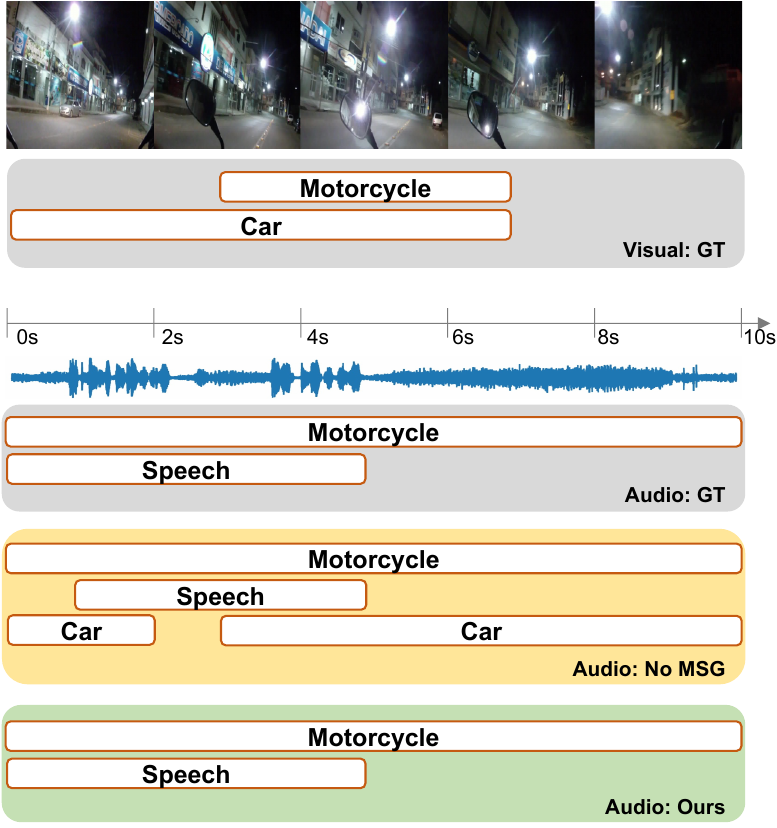}
\caption{Qualitative comparison of the messengers. `Visual' and `Audio' represent the visual and audio event, respectively. `GT' denotes ground truth. 
`No MSG' denotes model performs cross-modal fusion without messengers. `Ours' denotes the fusion with messengers.}
\label{messenger-fig}
\end{figure}

\subsection{Ablation Study}
\noindent\textbf{Effectiveness of the Messengers.}
Tab.~\ref{ablation messenger} shows the effectiveness of the messengers. 
`No Messenger' abbreviated as `No MSG' is the model which directly performs cross-modal fusion using Eqn.~\ref{fuse without messenger}. Compared with our final model `$n_\text{a}=1, n_\text{v}=1$', both the audio and visual performance decrease. We also provide qualitative analysis of the messenger in Fig.~\ref{messenger-fig}. 
`No MSG' wrongly detects the visible but not audible event `Car' on the audio stream due to the unconstrained visual context. 
By constraining the full visual context into our compact messenger, the irrelevant visual information `Car' is suppressed and only keeps the useful information of the `Motorcycle' for the audio.

In addition, we compare our messenger with fusion bottleneck token \cite{nagrani2021attention}, denoted as `MBT', by replacing the messenger with the same number of fusion bottleneck tokens in the transformer model. Our messenger `$n_\text{a}=1, n_\text{v}=1$' consistently outperforms `MBT' in the shallow transformer model, where only one layer of the encoder is used for each modality.

\vspace{3mm}
\noindent\textbf{Analysis of the number of messengers.}
Tab.~\ref{ablation messenger} also shows the performance of our model with different numbers of messengers. 
$n_\text{a}$ and $n_\text{v}$ is the number of audio and visual messengers, respectively. 
Using a large number of messengers shows decreasing performance, suggesting the dilution of the beneficial cross-modal context. 
Hence we adopt `$n_\text{a}=1, n_\text{v}=1$' in our final model.

\begin{figure*}[t]
\centering
\includegraphics[scale=0.62]{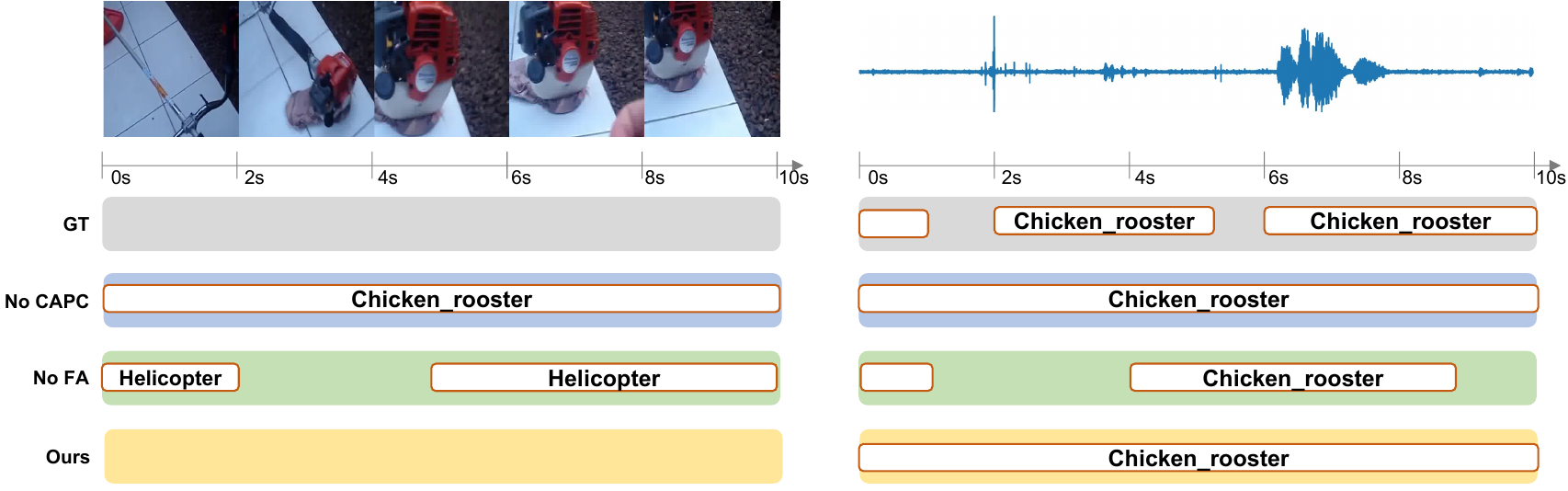}
\vspace{-3mm}
\caption{Qualitative comparison of CAPC. `GT' denotes the ground truth event labels. `No CAPC' denotes the model without cross-audio prediction consistency. `No FA' denotes the model without fusion with audio at the visual stream.}
\label{qualitative-result}
\end{figure*}

\begin{table}[t]
\centering
\resizebox{\linewidth}{!}{
\begin{tabular}{c|c|c|c|c|c}
\toprule
 & Audio & Visual & Audio-Visual & Type@AV & Event@AV \\
\midrule
No FA & \textbf{62.1} & 62.5  & 56.3  & 60.3  & 60.3 \\
No CAPC & 61.9 & 64.2  & 56.4  & 60.8  & 60.7  \\
Ours & 61.9  & \textbf{64.8} & \textbf{57.6} & \textbf{61.4} & \textbf{60.9} \\
\bottomrule
\end{tabular}
}
\vspace{-3mm}
\caption{Ablation study of the cross-audio prediction consistency. `No FA' denotes the model without fusion with audio at the visual stream. `No CAPC' denotes the model without using cross-audio prediction consistency. Segment-level results are reported}
\label{ablation consistency}
\end{table}

\vspace{3mm}
\noindent\textbf{Effectiveness of Cross-audio Prediction Consistency.}
Tab.~\ref{ablation consistency} presents the ablation study of cross-audio prediction consistency. `No CAPC' is the model trained without cross-audio prediction consistency. By forcing prediction consistency on the visual stream, \ie `Ours', both the performance of visual and audio-visual event detection show obvious improvement. 

We also verify whether CAPC leads to the trivial solution, \ie the visual prediction does not need the audio information at all. We replace the input to the cross-attention at the visual stream with the visual modality itself, \ie replacing $M_\text{a}$ with $S_\text{v}^L$ in Eqn.~\ref{eqn_fusion_messenger}, denoted as `No FA'. The performance is worse than the `No CAPC', and much worse than our full model `Ours'. We also provide the qualitative comparison in Fig.~\ref{qualitative-result}. `No FA' shows the poor generalization ability on the visual event detection as it wrongly detects a totally irrelevant event `Helicopter'. By learning with both its paired audio and other non-paired audios, the model can correctly identify that the `Chicken rooster' is an audible but not a visible event. 
This shows that our cross-audio prediction consistency does not de-activate the cross-modal fusion. Instead, it improves the robustness of the fusion between two non-fully correlated modalities.

\begin{table}[t]
\centering
\resizebox{\linewidth}{!}{
\begin{tabular}{c|c|c|c|c|c}
\toprule
$\mu$   & Audio & Visual & Audio-Visual & Type@AV & Event@AV \\
\midrule
0.1 & 61.3 & 63.6 & 56.3 & 60.4 & 60.1 \\
0.5 & \textbf{61.9}  & \textbf{64.8} & \textbf{57.6} & \textbf{61.4} & \textbf{60.9}\\
1 & \textbf{61.9} & 64.6 & \textbf{57.6} & \textbf{61.4} & 60.7  \\
\bottomrule
\end{tabular}
}
\vspace{-3mm}
\caption{Analysis of different value of $\mu$. Segment-level results are reported.}
\label{ablation mu}
\vspace{-3mm}
\end{table}

\vspace{3mm}
\noindent\textbf{Analysis of CAPC loss weight $\mu$.}
Tab.~\ref{ablation mu} presents the ablation study on different value of $\mu$.
Interestingly, using a small weight, \ie $\mu=0.1$ is worse than the model without CAPC (`No CAPC' in Tab.~\ref{ablation consistency}). The possible reason is that the model trained with `$\mu=0.1$' tends to focus on the easy training samples ($\text{A}_\text{orig}$ and $\text{A}_\text{rand}^i$ are both fully correlated with $\text{V}$). CAPC thus encourages V to take in full audio context to achieve faster convergence in this case, which provides the false signal. Only using larger weight can effectively optimize hard samples ($\text{A}_\text{orig}$ and $\text{A}_\text{rand}^i$ are not fully correlated with V), where CAPC guides visual stream to only select useful audio information. It can also be verified that CAPC loss of $\mu=0.1$ is much larger than $\mu=0.5$.
Therefore, we choose $\mu=0.5$ as our final setting.

\begin{table}[t]
\centering
\resizebox{\linewidth}{!}{
\begin{tabular}{c|c|c|c|c|c}
\toprule
N   & Audio & Visual & Audio-Visual & Type@AV & Event@AV \\
\midrule
1 & \textbf{61.9}  & \textbf{64.8} & \textbf{57.6} & \textbf{61.4} & \textbf{60.9}\\
2 & 61.7 & 64.5 & 57.0 & 61.1 & 60.5\\
3 & 61.7 & 64.0 &57.0 & 60.9 & 60.6  \\
\bottomrule
\end{tabular}
}
\vspace{-3mm}
\caption{Analysis of different number of random pairs N. Segment-level results are reported.}
\label{ablation N}
\vspace{-1mm}
\end{table}

\vspace{3mm}
\noindent\textbf{Analysis of the number of random pairs in CAPC.}
Tab.~\ref{ablation N} presents the ablation study of the number of random pairs in CAPC. 
Using a larger number of pairs, \ie, $N=3$, not only increases computation cost but also exhibits a decline in performance. 
We postulate the reason is that larger N provides a false signal that the visual modality should ignore any audio context (including correlated audio information). More analysis is provided in Supplementary Material. Therefore, we choose $N=1$ in our final model.

\section{Conclusion}
\label{conclusion}
We address the problem of fusion between two non-fully correlated modalities in weakly supervised audio-visual video parsing.
We propose the messenger-guided mid-fusion transformer to reduce the unnecessary cross-modal entanglement. The messengers act as fusion bottlenecks to mitigate the detrimental effect of the noisy labels.
Further, we propose cross-audio prediction consistency to reduce the negative interference of unmatched audio context to the visual stream. 
The effectiveness of our proposed method is analyzed both quantitatively and qualitatively.

\paragraph{Acknowledgement.} This research is supported by the National Research Foundation, Singapore under its AI Singapore Programme (AISG Award No: AISG2-RP-2021-024), and the
Tier 2 grant MOE-T2EP20120-0011 from the Singapore Ministry of Education.

{\small
\bibliographystyle{ieee_fullname}
\bibliography{PaperForReview}
}

\vspace{5mm}
\appendix
\renewcommand\thefigure{\Alph{section}\arabic{figure}}  
\renewcommand\thetable{\Alph{section}\arabic{table}}

{\centering\section*{Supplementary Material}}

\section{Additional Analysis of CAPC.}
\setcounter{figure}{0}
\setcounter{table}{0}
Tab.~\ref{CAPC} presents the results of detecting visual events by different models. 
We split the visual events into visual-exclusive and multi-modality events. Visual-exclusive events refers to events only happening in the visual modality, while multi-modality events appear on audio and visual streams with temporal overlap. The results are averaged F-scores per event.
As shown in Tab.~\ref{CAPC}, the results for the visual-exclusive events consistently improve with an increase in the value of $N$, suggesting the CAPC can effectively reduce the influence of unmatched audio context. 
However, the accuracy of multi-modality events drops when $N=3$, suggesting that large $N$ can impede the positive impact of audio on its correlated visual event. Only a small number of $N$, \ie $N=1$, improves fusion effectiveness for both events.

\begin{table}[h]
\centering
\resizebox{0.7\linewidth}{!}{
\begin{tabular}{c|cc}
\hline
\multirow{2}{*}{Model} & \multicolumn{2}{c}{Visual Event} \\ \cline{2-3} 
                       & Visual-exclusive       & Multi-modality      \\ \hline
No CAPC                & 35.6              & 63.0              \\
N=1                    & 38.5              & \textbf{63.6}              \\
N=3                    & \textbf{38.6}              & 62.9              \\ \hline
\end{tabular}
}
\vspace{-3mm}
\caption{Additional Analysis of cross-audio prediction consistency (CAPC).}
\label{CAPC}
\end{table}

\section{Qualitative Comparison with HAN}
\begin{figure*}[t]
\centering
\includegraphics[scale=0.62]{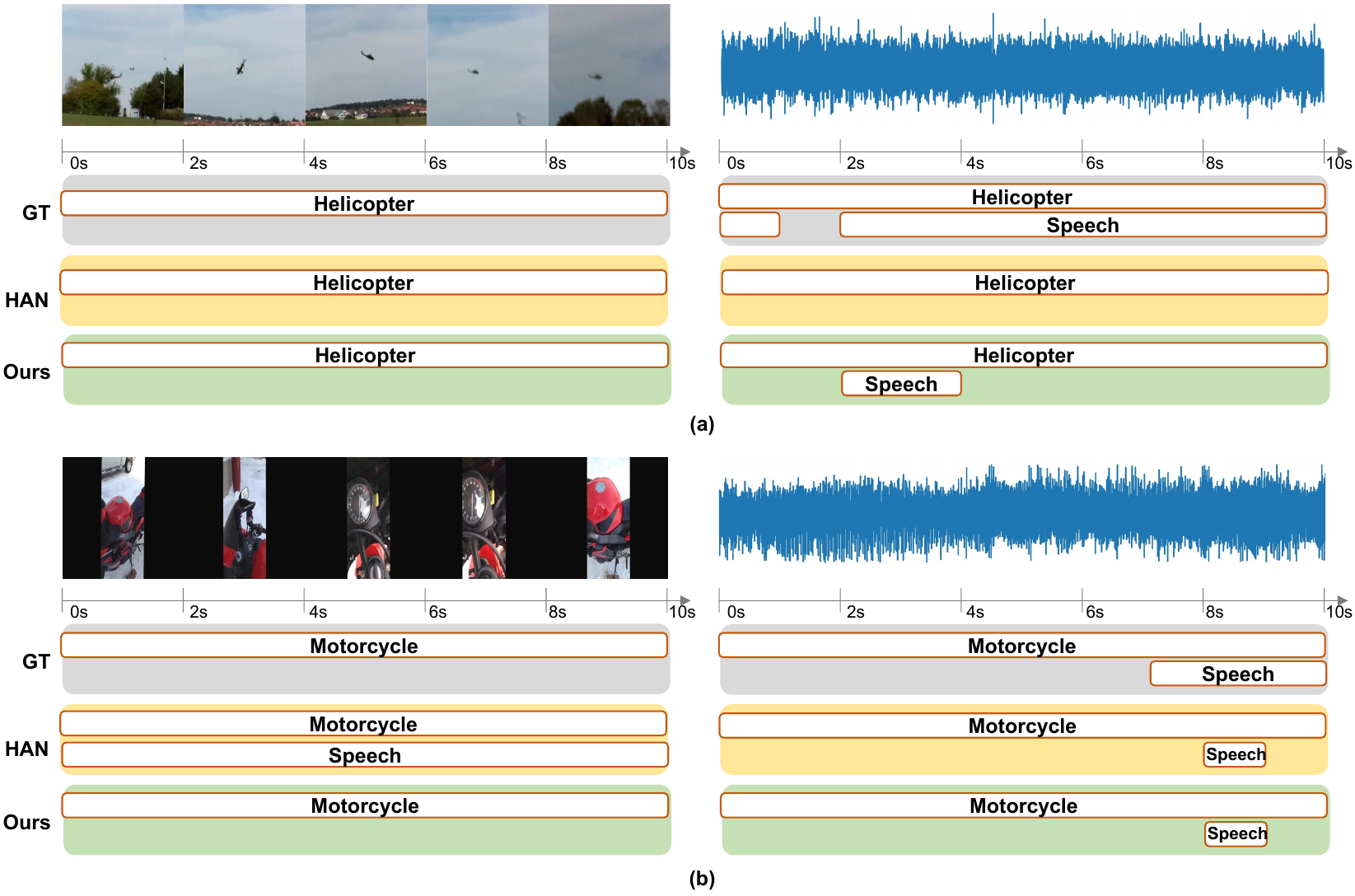}
\caption{Qualitative comparison with HAN.}
\label{compare-han}
\end{figure*}

Fig. ~\ref{compare-han} shows the qualitative results. 
HAN \cite{tian2020unified} fails to identify the single-modality event ('Speech') in Fig. ~\ref{compare-han}(a) or wrongly detect it on two modalities in Fig. ~\ref{compare-han}(b), suggesting the information on the audio and visual streams are highly confounded.  In contrast, our model can correctly detect the audio and visual events.

\section{\ytComment{Illustration of Single-modality and Multi-modality Event}}
Single-modality event in Table 1 of main paper refers to events \textit{only} happening in one modality, while audio (visual) event in Table 2 includes event both only happening in the audio (visual) and happening in the audio and visual modality. 

Multi-modality event in Table 1 of main paper refers to event appearing with temporal overlap (either partial or full overlap) on audio and visual streams, while audio-visual event in Table 2 refers to event with \textit{full} overlap on audio and visual streams. 

The split of single-modality and multi-modality event in Table 1 is to illustrate that the strong entanglement with another non-fully correlated modality is harmful in detecting its own exclusive events. 
The evaluation of audio, visual and audio-visual event in Table 2 is the standard benchmark of audio-visual video parsing task \cite{tian2020unified}.

\end{document}